\documentclass[runningheads]{llncs}
\usepackage[T1]{fontenc}
\usepackage{graphicx}
\usepackage{booktabs}
\usepackage[misc]{ifsym}

\usepackage{mwe}

\usepackage{amsmath} 
\usepackage{amssymb} 
\usepackage{subfigure}
\usepackage{multirow}
\usepackage{tikz} 

\begin{document}

\title{Not All Preferences are What You Need for Post-Training: Selective Alignment Strategy for Preference Optimization}


\titlerunning{Not All Prefrences are What You Need for Post-Training}

\author{Zhijin Dong}

\institute{Peking University \\ \email{zhijindong@pku.edu.cn}}

\maketitle              

\begin{abstract}
Post-training alignment of large language models (LLMs) is a critical challenge, as not all tokens contribute equally to model performance. This paper introduces a selective alignment strategy that prioritizes high-impact tokens within preference pairs, leveraging token-level log-probability differences between the current policy and a reference model. By focusing on these informative tokens, our approach reduces computational overhead and enhances alignment fidelity. We further explore the role of reference model quality, demonstrating that stronger reference models significantly improve token selection accuracy and overall optimization effectiveness. Comprehensive experiments on benchmarks such as Arena-Hard and MT-Bench validate the superiority of our Selective-DPO method over standard DPO and distillation-based baselines. Our findings highlight the importance of token-level optimization and reference model selection in advancing preference alignment for LLMs. 
The code is available at \url{https://github.com/Dongzhijin/SDPO}.

\end{abstract}

\section{Introduction}

Large language models (LLMs) have revolutionized natural language processing, enabling applications ranging from conversational agents to code generation. Despite their remarkable capabilities, aligning LLMs with human preferences remains a critical challenge, particularly in the post-training stage. Effective alignment ensures that models not only produce fluent text but also adhere to nuanced human values and expectations.

Traditional approaches to preference alignment, such as Reinforcement Learning from Human Feedback (RLHF), rely on algorithms like Proximal Policy Optimization (PPO)\cite{ppo2017schulman}. While effective, these methods introduce significant computational overhead and instability due to the need for reward model training and repeated policy sampling. Direct Preference Optimization (DPO) \cite{rafailov2024directpreferenceoptimizationlanguage} has emerged as a promising alternative, offering a more efficient framework by directly optimizing model parameters using preference pairs without explicit reinforcement learning. Variants of DPO, such as SimPO and Step-DPO, have demonstrated improved stability and efficiency in aligning LLMs with human feedback.

Recent research\cite{sepo2024tokenlevel,lin2025rho1tokensneed} highlights that not all tokens in a sequence contribute equally to preference alignment. Token-level selective optimization methods, such as SePO, focus on identifying and optimizing the most informative tokens, thereby improving training efficiency and generalization. However, existing approaches often rely on complex reward estimation or oracle models, which may limit their practicality.

In this work, we propose a novel selective alignment strategy for preference optimization. Our method leverages token-level log-probability differences between the current policy and a reference model to score and select the most critical tokens for optimization. By focusing on high-impact tokens, this approach reduces computational overhead and enhances alignment fidelity. Additionally, we explore the role of reference model quality in improving token selection accuracy and overall alignment performance.

Through comprehensive experiments on challenging benchmarks such as Arena-Hard and MT-Bench, we demonstrate that our selective alignment approach achieves superior alignment quality and training efficiency compared to standard DPO and distillation-based baselines. Our findings underscore the importance of focusing on informative tokens and leveraging high-quality reference models in preference optimization, paving the way for more efficient and effective alignment methods.

\section{Related Work}

\subsection{Direct Preference Optimization}

Aligning large language models (LLMs) with human preferences has become a central challenge in post-training. A common paradigm for this task is reinforcement learning from human feedback (RLHF), where Proximal Policy Optimization (PPO) \cite{ppo2017schulman} has been widely adopted. PPO alternates between sampling from the model and optimizing it using a learned reward model. However, PPO introduces substantial computational overhead and often suffers from training instability.

To address these limitations, Direct Preference Optimization (DPO) \cite{rafailov2024directpreferenceoptimizationlanguage}was proposed as a more efficient alternative. DPO eliminates the need for an explicit reward model or reinforcement learning by directly optimizing the model using preference pairs. It adopts the Bradley-Terry model \cite{bradley1952rank} to model pairwise preferences and reparameterizes the reward function using the log-probability ratio between the current policy and a reference policy. This leads to a simple classification loss that is both stable and efficient.

Several extensions to DPO have been developed to handle specific challenges. SimPO \cite{meng2024simposimplepreferenceoptimization} simplifies the reward computation by removing the reference model and normalizing the sequence log-probability, while Step-DPO \cite{lai2024stepdpostepwisepreferenceoptimization} improves performance in long-form reasoning tasks by applying DPO at the step level rather than the sequence level.

Our method builds upon this line of work, and a detailed derivation of DPO and its extensions is presented in Section~\ref{sec:Preliminaries} to facilitate our approach.

\subsection{Direct Preference Knowledge Distillation}

Recent studies have explored integrating preference supervision into the knowledge distillation pipeline. Direct Preference Knowledge Distillation (DPKD) \cite{li2025directpreferenceknowledgedistillation} leverages human preference signals as implicit rewards during the distillation process from a larger teacher to a smaller student model. In contrast to traditional distillation that relies solely on token-level KL divergence, DPKD incorporates preference-based objectives to guide the student's learning.

DPKD typically operates in two stages: the teacher is first aligned using preference optimization techniques (such as DPO), and then the student is trained to match both its outputs and its preference-aware behaviors. This two-stage process enables the student to inherit not only factual knowledge but also alignment properties, improving performance particularly in alignment-sensitive tasks. DPKD has shown improved sample quality and alignment fidelity compared to standard distillation baselines.

\subsection{Token-Level Selective Optimization}

Recent studies have shown that not all tokens are equally informative during training. Lin et al.~\cite{lin2025rho1tokensneed} propose a method for efficient pretraining by assigning weights to tokens based on their estimated learning utility. Their approach selectively emphasizes high-value tokens to reduce redundant computation and improve generalization. Although developed in the context of language model pretraining, this principle naturally extends to the post-training alignment stage.

SePO~\cite{sepo2024tokenlevel} applies this idea to preference optimization. It builds upon DPO by estimating token-level rewards using an oracle model and selecting only the most informative tokens from preference pairs for optimization. SePO models the generation process as a token-level MDP and assumes that the total reward can be decomposed additively along the sequence. By optimizing only high-scoring tokens from preferred responses and low-scoring tokens from rejected ones, SePO achieves improved training efficiency without compromising alignment quality.

Our work is closely related in motivation but diverges in its implementation. Instead of relying on explicit reward estimation from an oracle model, we propose a simpler scoring strategy based on log-probability differences between the current policy and a reference model. Moreover, we explore how using stronger reference models improves token scoring fidelity and leads to more effective selective optimization.

\section{Preliminaries}
\label{sec:Preliminaries}

Proximal Policy Optimization (PPO) \cite{ppo2017schulman} is a standard reinforcement learning algorithm used in reinforcement learning from human feedback (RLHF). It updates the policy \(\pi_\theta\) using a learned reward model \(r_\phi(x, y)\), with the objective:
\begin{equation}
\max_{\pi_\theta} \mathbb{E}_{x \sim D, y \sim \pi_\theta} \left[ r_\phi(x, y) \right] - \beta D_{\text{KL}}[\pi_\theta(y|x) || \pi_{\text{ref}}(y|x)],
\end{equation}
where \(\pi_{\text{ref}}\) is the reference policy and \(\beta\) is a KL regularization coefficient.

While PPO is effective, it suffers from high computational cost due to repeated reward model training and policy sampling, and is often unstable.

To address these challenges, Direct Preference Optimization (DPO) was proposed. DPO directly optimizes the policy from preference data without reinforcement learning. Given a dataset \( D = \{(x, y^w, y^l)\} \), DPO models preference using the Bradley-Terry framework:
\begin{equation}
p(y^w \succ y^l | x) = \frac{\exp(r(x, y^w))}{\exp(r(x, y^w)) + \exp(r(x, y^l))},
\end{equation}
where the reward is reparameterized as:
\begin{equation}
r(x, y) = \beta \log \frac{\pi_\theta(y|x)}{\pi_{\text{ref}}(y|x)} + \beta \log Z(x),
\end{equation}
leading to the final loss:
\begin{equation}
L_{\text{DPO}} = - \mathbb{E}_{(x, y^w, y^l)} \left[ \log \sigma \left( \beta \log \frac{\pi_\theta(y^w|x)}{\pi_{\text{ref}}(y^w|x)} - \beta \log \frac{\pi_\theta(y^l|x)}{\pi_{\text{ref}}(y^l|x)} \right) \right].
\end{equation}

DPO simplifies training by removing reward modeling and sampling, offering improved stability and efficiency. Variants like SimPO and Step-DPO further adapt this framework for response length normalization and reasoning tasks, respectively.

Building upon these advancements, our work introduces a selective alignment strategy that leverages token-level log-probability differences to identify and optimize high-impact tokens. This approach aims to address inefficiencies in sequence-level alignment and improve the fidelity of preference optimization, as detailed in the following section.

\section{Method}  
\label{sec:Method}
\begin{figure}[t]
\centering
\begin{tikzpicture}[node distance=1.5cm, auto]

\node[rectangle, draw, text centered, minimum width=4cm, minimum height=1cm] (input) {Input: Preference Dataset, Policy Model, Reference Model};
\node[rectangle, draw, text centered, minimum width=4cm, minimum height=1cm, below of=input] (score) {Step 1: Compute Alignment Scores};
\node[rectangle, draw, text centered, minimum width=4cm, minimum height=1cm, below of=score] (select) {Step 2: Select Top \(k\%\) High-Impact Tokens};
\node[rectangle, draw, text centered, minimum width=4cm, minimum height=1cm, below of=select] (optimize) {Step 3: Optimize Policy Using Selective-DPO Loss};
\node[rectangle, draw, text centered, minimum width=4cm, minimum height=1cm, below of=optimize] (output) {Output: Optimized Policy Model};

\draw[->] (input) -- (score);
\draw[->] (score) -- (select);
\draw[->] (select) -- (optimize);
\draw[->] (optimize) -- (output);

\end{tikzpicture}
\caption{Workflow of the Selective Alignment Strategy for Preference Optimization.
  The method focuses on high-impact tokens within preference pairs, leveraging token-level log-probability differences to score and select the most informative tokens for optimization. The process consists of three main steps: computing alignment scores, selecting top \(k\%\) high-impact tokens, and optimizing the policy using selective-DPO loss.}
\label{fig:method_workflow}
\end{figure}
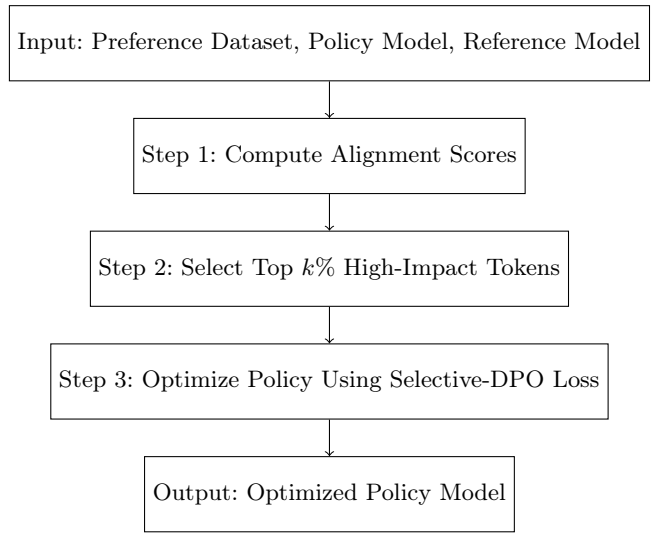

In this section, we introduce our selective alignment strategy for preference optimization, which focuses on high-impact tokens within preference pairs. The key idea is to leverage token-level log-probability differences between the current policy and a reference model to score and select the most informative tokens for optimization. This approach aims to improve training efficiency and alignment quality by focusing on tokens that carry significant information regarding user preferences. 
The pipeline of our selective alignment strategy is illustrated in Figure~\ref{fig:method_workflow}. The method consists of three main steps: computing alignment scores, selecting top \(k\%\) high-impact tokens, and optimizing the policy using selective-DPO loss.

\subsection{Selective Alignment in Preference Optimization}

Post-training alignment of large language models (LLMs) requires recognizing that not all tokens contribute equally to model performance. Certain tokens carry significant information regarding user preferences, while others may be less relevant or even detrimental. Addressing this disparity is crucial for enhancing the efficiency and effectiveness of the training process. To tackle this challenge, we propose a selective alignment strategy that focuses on optimizing high-impact tokens within preference pairs. Our selective alignment strategy leverages token-level log-probability differences between the current policy and a reference model to score and select the most impactful tokens for optimization. By focusing on these high-impact tokens, we aim to improve training efficiency and avoid overfitting to irrelevant parts.

We begin with the token-level log-likelihood:
\begin{equation}
\log \pi(y \mid x) = \sum_{i=1}^{|y|} \log \pi(y_i \mid x, y_{<i}),
\end{equation}
where \( |y| \) denotes the sequence length, and \( y_{<i} \) is the prefix up to the \( i \)-th token.  

To evaluate token importance, we define an alignment score:
\begin{equation}
s(y_{i}) = (-1)^{ \mathbb{I}(y_{i} \in y^l)} \cdot [\log \pi_{\text{ref}}(y_{i} \mid x, y_{<i}) - \log \pi_\theta(y_{i} \mid x, y_{<i})].
\end{equation}
where \( \mathbb{I}(y_{i} \in y^l) \) indicates whether \( y_i \) belongs to the "lose" sequence.

This score captures discrepancies between the reference model \(\pi_{\text{ref}}\) and the current policy \(\pi_\theta\), emphasizing tokens that require alignment. By selecting the top \( k\% \) of tokens based on their scores, we focus optimization on the most informative parts, improving efficiency and alignment quality.

To explain the alignment score, we leverage the complementary roles of \(\pi_{\text{ref}}\) and \(\pi_\theta\):
\begin{itemize}
  \item \(\pi_{\text{ref}}\), a stronger reference model, serves as a guide. For the "win" sequence (\(y^w\)), it assigns higher token-level log-likelihoods to tokens deemed critical for alignment, indicating areas requiring optimization. Conversely, for the "lose" sequence (\(y^l\)), it assigns lower log-likelihoods, suggesting tokens that should be de-emphasized.
  \item \(\pi_\theta\), the model being optimized, reflects its current alignment state. For the "win" sequence, higher token-level log-likelihoods indicate tokens that are already well-aligned, while for the "lose" sequence, higher log-likelihoods highlight tokens that are misaligned.
\end{itemize}

By combining these perspectives, the alignment score identifies tokens that \(\pi_{\text{ref}}\) considers important for optimization and that \(\pi_\theta\) has yet to align properly. This ensures that the optimization process focuses on the most impactful tokens, improving efficiency and alignment quality.

Tokens with the highest alignment scores are selected for optimization. Specifically, the top \( k\% \) of tokens ranked by their scores are chosen, reducing noise and focusing on the most informative parts. The Selective-DPO loss is then defined as:
\begin{equation}
  L_{\text{Selective-DPO}} = - \mathbb{E}_{(x, y^w, y^l) \sim D} \left[ \log \sigma\left(\beta R(y^w)-\beta R(y^l)  \right) \right]
  \end{equation}
where \( R(y) \) is the selective alignment reward function.
\begin{equation}
  R(y)=\sum_{i=1}^{|y|} \mathbb{I}(s(y_{i}) \in topk\%) \cdot [ \log \frac{\pi_{\theta}(y_i \mid x, y_{<i})}{\pi_{ref}(y_i \mid x, y_{<i})} ]
\end{equation}
where \( topk\% \) denotes the top \( k\% \) of tokens ranked by their alignment scores.

\subsection{Reference Model Selection and Knowledge Distillation}

The choice of reference model is pivotal in our selective alignment approach, as it directly impacts the accuracy of token selection and the overall alignment performance. A high-quality reference model serves as a teacher, guiding the optimization of the policy and enhancing the alignment scores for more effective token identification.

A stronger reference model, such as a larger model, typically offers superior performance due to scaling laws. Larger models provide more accurate alignment scores, which improve token selection fidelity and lead to better optimization outcomes. However, the increased training cost associated with larger models must be considered.

Alternatively, a model of the same size that has undergone DPO alignment training on the same dataset can be used as the reference model. Such a model combines the advantages of alignment-specific training with computational efficiency, offering a practical solution for refining the alignment process. By leveraging the alignment properties of a pre-trained reference model, the optimization process can achieve higher accuracy in token selection and improved overall alignment quality.

In essence, the reference model acts as a knowledge distillation source, similar to Direct Preference Knowledge Distillation (DPKD) \cite{li2025directpreferenceknowledgedistillation}, guiding the policy optimization through Selective-DPO loss.   
This dual role of the reference model—as both a performance enhancer and a computationally efficient alternative—underscores its importance in achieving effective selective alignment.

\section{Experiments}
\subsection{Experimental Setup}

\paragraph{Models.}
We conduct experiments using two models: a 0.5-billion-parameter model (denoted as \textit{0.5B-SFT}) and a 3-billion-parameter model (denoted as \textit{3B-SFT}). To ensure model confidentiality, actual model names are anonymized. Both models are based on the Transformer architecture and serve as the backbone for our preference optimization experiments. The models are initialized from pre-trained checkpoints and fine-tuned using SFT data. For reference models, we adopt multiple configurations to evaluate the impact of reference quality on alignment performance. These include using the base model itself as the reference, a same-size model aligned via DPO, and a larger (stronger) model trained via DPO.

\paragraph{Baselines.}
To evaluate the effectiveness of our proposed methods, we compare them against the following baselines:
\begin{itemize}
  \item \textbf{Supervised Fine-Tuning (SFT)}: This serves as the simplest baseline, where the model is fine-tuned using supervised learning on labeled data without preference optimization.
  \item \textbf{Direct Preference Optimization (DPO)}: This is the main baseline, providing a direct approach for optimizing the model based on human preference data without using reinforcement learning. DPO eliminates the need for explicit reward modeling and sampling, offering a stable and efficient alternative to RLHF.
  \item \textbf{Direct Preference Knowledge Distillation (Distill-DPO)}: This method integrates preference supervision into the knowledge distillation pipeline. It distills the alignment properties of a teacher model, fine-tuned using preference data, into a student model, enabling us to evaluate the impact of preference-aware knowledge transfer on alignment performance.
\end{itemize}

\paragraph{Datasets.}
We use the Skywork-Reward-Preference-80K-v0.2 dataset~\footnote{\url{https://huggingface.co/datasets/Skywork/Skywork-Reward-Preference-80K-v0.2}}, a large-scale Chinese preference dataset with 80,000 human-annotated pairs. Each sample includes a prompt, two responses, and a preference label. The dataset covers diverse domains like math, code, safety, and reasoning, making it suitable for evaluating preference optimization. For details, see~\cite{liu2024skywork}.

\paragraph{Evaluation Benchmarks.}  
To comprehensively assess our model alignment strategies, we utilize two widely recognized benchmarks: \textbf{Arena-Hard:} A challenging evaluation suite based on 500 user queries sourced from the Chatbot Arena platform. These queries are specifically chosen for their complexity, requiring nuanced reasoning, ethical decision-making, and robust common-sense knowledge. The benchmark uses GPT-4-Turbo as a reviewer to score responses by comparing them against a baseline model (e.g., GPT-4-0314). This benchmark is designed to stress-test a model’s capacity to handle intricate scenarios while closely aligning with human preferences \cite{arena2023benchmark}. \textbf{MT-Bench:} A benchmark that focuses on multi-turn dialogue evaluation. It contains 80 carefully crafted question sets that span a diverse range of categories, including writing, role-playing, information extraction, reasoning, mathematics, coding, and STEM as well as humanities knowledge. By assessing how well models maintain relevance, consistency, and coherence across multiple interactions, MT-Bench provides valuable insights into their conversational alignment performance \cite{mtbench2023evaluation}.

\paragraph{Training Details}
Training is implemented using the TRL \texttt{DPOTrainer} with DeepSpeed ZeRO-3 optimization and bf16 precision. We use a per-device batch size of 1 and a global batch size of 1280 achieved via gradient accumulation. The training is conducted for 2 epochs with a maximum input length of 4096 tokens. For Selective-DPO, we set the KL regularization coefficient \(\beta = 0.01\), and retain the top 40\% of tokens based on alignment scores computed from token-level log-likelihood differences between the policy and reference model. These hyperparameters are selected based on ablation studies and preliminary tuning.

\subsection{Main Results}

\begin{table*}[t]
  \centering
  \caption{Comparison of different alignment methods on Arena-Hard (win rate, \%) and MT-Bench (total score) for 0.5B and 3B models.}
  \begin{tabular}{lcccccc}
    \toprule
    \multirow{2}{*}{Method} & \multicolumn{3}{c}{0.5B-SFT} & \multicolumn{3}{c}{3B-SFT} \\
    \cmidrule(lr){2-4} \cmidrule(lr){5-7}
    & Ref Model & Arena-Hard (\%) & MT-Bench & Ref Model & Arena-Hard (\%) & MT-Bench \\
    \midrule
    SFT           & -             & 12.1     & 6.89     & -             & 15.3     & 7.35 \\
    DPO           & SFT           & 13.4     & 6.95     & SFT           & 16.6     & 7.43 \\
    Distill-DPO   & 10B-instruct  & 20.3     & 7.12     & 33B-instruct  & 24.7     & 7.67 \\
    Selective-DPO & 0.5B-instruct  & 14.5     & 7.04     & 3B-instruct  & 17.9     & 7.54 \\
    Selective-DPO & 10B-instruct  & 22.5     & 7.34     & 33B-instruct  & 26.9     & 7.93 \\
    \bottomrule
  \end{tabular}
  
  \label{tab:main_results}
\end{table*}

Table~\ref{tab:main_results} presents the performance comparison of different alignment methods on two benchmarks, Arena-Hard and MT-Bench, for both 0.5B and 3B models. The results demonstrate the effectiveness of our proposed Selective-DPO method in improving alignment quality and efficiency.

For the 0.5B-SFT model, Selective-DPO with a 10B-instruct reference model achieves a significant improvement in Arena-Hard win rate (22.5\%) and MT-Bench total score (7.34) compared to the baseline DPO (13.4\% and 6.95, respectively). This highlights the advantage of focusing on high-impact tokens during optimization. Similarly, for the 3B-SFT model, Selective-DPO with a 33B-instruct reference model achieves the best performance, with a 26.9\% win rate on Arena-Hard and a 7.93 total score on MT-Bench, outperforming both DPO and Distill-DPO.

The results also show that using stronger reference models (e.g., 10B-instruct and 33B-instruct) further enhances the effectiveness of Selective-DPO. This is evident from the consistent performance gains over weaker reference models (e.g., 0.5B-instruct and 3B-instruct). These findings validate our hypothesis that stronger reference models provide more accurate token-level alignment scores, leading to better optimization outcomes.

Compared to Distill-DPO, which also leverages larger reference models, Selective-DPO achieves comparable or superior performance while maintaining a simpler optimization framework. For instance, Selective-DPO with a 33B-instruct reference model achieves a higher Arena-Hard win rate (26.9\% vs. 24.7\%) and MT-Bench score (7.93 vs. 7.67) for the 3B-SFT model.

Overall, the results demonstrate that Selective-DPO effectively identifies and optimizes high-impact tokens, leading to improved alignment quality and efficiency across different model sizes and benchmarks.

\subsection{Ablation Studies: Effect of Token Selection Ratio}

To investigate the impact of different token selection ratios on our selective alignment approach, we conduct experiments with varying percentages of selected tokens using the 0.5B-SFT model and 10B-instruct reference model. Table~\ref{tab:topk_ablation} presents the results.

\begin{table}[t]
  \centering
  \caption{Performance with different token selection ratios (0.5B-SFT model, 10B-instruct reference).}
  \begin{tabular}{lcc}
    \toprule
    Top $k\%$ & Arena-Hard (\%) & MT-Bench \\
    \midrule
    20\%        & 18.7            & 7.12     \\
    40\%        & \textbf{22.5}   & \textbf{7.34} \\
    60\%        & 20.9            & 7.21     \\
    80\%        & 19.3            & 7.15     \\
    \bottomrule
  \end{tabular}
  \label{tab:topk_ablation}
\end{table}

The results demonstrate that selecting the top 40\% of tokens achieves optimal performance across both evaluation metrics. This indicates that focusing on a moderate proportion of high-impact tokens effectively captures preference-relevant information while excluding potentially noisy tokens. The performance degradation observed at both lower (20\%) and higher (60\%, 80\%) selection ratios confirms our hypothesis that neither too few nor too many tokens are optimal for alignment. Too few tokens may miss important information, while too many may introduce noise from less relevant tokens, diluting the optimization focus.

\subsection{Hyperparameter Analysis: Effect of Regularization Coefficient}

We investigate the impact of different values for the regularization coefficient $\beta$ on model performance. This parameter controls the strength of KL regularization between the policy and reference model. Table~\ref{tab:beta_ablation} shows the results for Selective-DPO with the 0.5B-SFT model and 10B-instruct reference model across different $\beta$ values.

\begin{table}[t]
  \centering
  \caption{Performance with different regularization coefficients $\beta$ (0.5B-SFT model, 10B-instruct reference).}
  \begin{tabular}{lcc}
    \toprule
    $\beta$ value & Arena-Hard (\%) & MT-Bench \\
    \midrule
    0.001       & 18.6            & 7.15     \\
    0.01        & \textbf{22.5}   & \textbf{7.34} \\
    0.02        & 21.8            & 7.28     \\
    0.05        & 19.7            & 7.19     \\
    0.1         & 17.9            & 7.06     \\
    \bottomrule
  \end{tabular}
  \label{tab:beta_ablation}
\end{table}

The results indicate that $\beta = 0.01$ achieves the best performance across both evaluation metrics for the 0.5B-SFT model. Too small values (e.g., 0.001) result in insufficient regularization, allowing the model to deviate too much from the reference policy. Conversely, larger values (e.g., 0.05, 0.1) impose excessive constraints, limiting the model's ability to learn from preference data. The moderate value of $\beta = 0.01$ strikes the optimal balance between learning from preferences and maintaining the stability provided by the reference model.

\section{Discussion}
\subsection{Limitations}
While our selective alignment approach demonstrates significant improvements in preference optimization, it is not without limitations. One key limitation is the reliance on a reference model to compute token-level alignment scores. The effectiveness of our method is contingent on the quality and relevance of the reference model used. If the reference model is not well-aligned or does not capture the necessary nuances of the task, it may lead to suboptimal token selection and alignment performance.

Additionally, our method currently focuses on token-level alignment without considering the broader context of the entire sequence. While this allows for efficient optimization, it may overlook interactions between tokens that could be important for certain tasks. Future work could explore incorporating contextual information into the token selection process to further enhance alignment quality.

Finally, our method shares a common limitation with DPO-based approaches (especially SimPO). Specifically, while these methods align well with subjective metrics as response style, they exhibit certain performance constraints on objective metrics like instruction-following. This limitation was observed during our experiments, where the IFEval metric showed a decline. We hypothesize that this is primarily due to an excessive focus on subjective preferences during optimization, which may weaken the model's generalization ability for objective tasks. Future work could explore specialized optimization strategies for instruction-following tasks to further improve the model's performance on objective metrics.

\section{Conclusion}
In this work, we proposed a selective alignment strategy for preference optimization in large language models. By focusing on high-impact tokens based on token-level log-probability differences, our method improves alignment quality compared to standard DPO and distillation-based approaches. We demonstrated the effectiveness of our approach through comprehensive experiments on challenging benchmarks, showing that selective alignment can significantly enhance model performance while reducing computational overhead. Our findings highlight the importance of token selection in preference optimization and open avenues for further research in efficient alignment methods.

\section*{Acknowledgments}
We would like to express our heartfelt gratitude to my advisor for their invaluable guidance and support throughout this research. We also extend our thanks to colleagues and collaborators for their insightful discussions and constructive feedback during the internship. Finally, we acknowledge the company for providing resources and support during the internship period, which greatly contributed to the success of this work.

%
%
\bibliographystyle{splncs04}
\bibliography{references}
%




\end{document}